\documentclass[times,10pt]{article}
\usepackage[letterpaper, total={6.5in, 9in}]{geometry}
\usepackage{graphicx}
\usepackage{amsmath,amssymb}
\newtheorem{theorem}{Theorem}
\newtheorem{corollary}{Corollary}

\title{Equivalence of stochastic and deterministic policy gradients}
\author{Emo Todorov\footnote{The author is with Roboti LLC and the University of Washington. Email: etodorov@gmail.com}}

\begin{document}

\maketitle

\begin{abstract}
Policy gradients in continuous control have been derived for both stochastic and deterministic policies. Here we study the relationship between the two. In a widely-used family of MDPs involving Gaussian control noise and quadratic control costs, we show that the stochastic and deterministic policy gradients, natural gradients, and state value functions are identical; while the state-control value functions are different. We then develop a general procedure for constructing an MDP with deterministic policy that is equivalent to a given MDP with stochastic policy. The controls of this new MDP are the sufficient statistics of the stochastic policy in the original MDP. Our results suggest that policy gradient methods can be unified by approximating state value functions rather than state-control value functions.
\end{abstract}

\section{Introduction}

Policy gradients methods \cite{williams1992,sutton1999,kakade2001,peters2008,silver2014} are now widely used, and have produced impressive empirical results. In continuous control, they have been derived for both stochastic \cite{sutton1999} and deterministic \cite{silver2014} policies. The resulting algorithms have different strengths and weaknesses, even though there are a number of similarities in how they are applied in practice, and recent generalizations \cite{ciosek2020,pfau2025} have pointed to even deeper similarities. Here we study the relationship between the two. First we focus on MDPs with Gaussian control noise and quadratic control cost (while the dynamics and state cost remain general) and show that deterministic and stochastic policy gradients for such MDPs are equivalent. We then develop a much more general result, where any MDP with stochastic policy can be converted into an equivalent MDP with deterministic policy. The new MDP has the same state and policy parameters, but a different control space -- namely the sufficient statistics of the stochastic policy in the original MDP. The only quantities that are not equivalent (in either the special or general case) are the state-control value functions. The latter observation suggests that policy gradient methods can be unified by approximating state value functions, instead of the common practice of approximating state-control value functions. We briefly outline the corresponding learning algorithms.

\section{Background: Policy gradients in continuous MDPs}

Let $x \in R^{n_x}$ and $u \in R^{n_u}$ be the state and control vectors of a continuous MDP with dynamics $x' \sim p(\cdot|x,u)$, initial state distribution $p_0(x)$, step cost $\ell(x,u)$ and discount factor $\gamma \in (0,1)$. We consider stochastic policies $u \sim \pi(\cdot|x,\theta)$ as well as deterministic policies $u = \mu(x, \theta)$, both parameterized by $\theta \in R^{n_\theta}$. The symbols (S) and (D) will be used to disambiguate between these two problem formulations when needed.

Define the state value function $v(x, \theta)$ as the total discounted cost starting at state $x$ and following policy $\pi$ or $\mu$ with parameters $\theta$:
\begin{equation}  
v(x, \theta) \equiv \mathbb{E}_{x_t, u_t} \left[ \sum\nolimits_{t = 0}^\infty {\gamma^t \ell\left( x_t, u_t \right) }\right]
\end{equation}
where $x_0 = x$, $x_{t+1} \sim p(\cdot | x_t, u_t )$, and $u_t \sim \pi(\cdot | x_t,\theta)$ or $u_t = \mu(x_t, \theta)$ respectively. This function satisfies the Bellman equation, which has slightly different form in the two formulations:
\begin{equation}
\begin{aligned}
v_S(x,\theta) & = \mathbb{E}_{u \sim \pi(\cdot | x, \theta)} \left[ Q_S(x,u,\theta) \right] \\
v_D(x,\theta) & =  Q_D(x,\mu(x, \theta),\theta) 
\end{aligned}
\end{equation}
Here $Q(x,u,\theta)$ is the state-control value function, defined in both formulations as:
\begin{equation}
Q(x,u,\theta) \equiv \ell(x, u) + \gamma \mathbb{E}_{x' \sim p(\cdot|x,u)} [v(x',\theta))]
\end{equation}

\noindent The performance $J(\theta)$ of the policy is then:
\begin{equation}
J(\theta) \equiv \mathbb{E}_{x \sim p_0(\cdot)} [ v(x, \theta) ]
\end{equation}
Policy gradient methods  seeks to compute $\nabla_\theta J$, and use it to improve the policy.

The above functions $v, Q$ refer to the exact value functions which depend on the policy parameters $\theta$; not to be confused with parametric function approximation. In the remainder of this paper we will often drop the explicit dependence of $v,Q$ on $\theta$ to avoid clutter.

Existing policy gradient results are in the form of expectations, which in episodic settings are taken with respect to a ``discounted density'':
\begin{equation}
\rho(x, \theta) \equiv \int p_0(y) \left( \sum\nolimits_{t = 1}^\infty { \gamma^{t-1} p(y \rightarrow x, t, \theta)} \right) dy
\end{equation}
Here $p(y \rightarrow x, t, \theta)$ is the probability of transitioning from state $y$ to state $x$ after $t$ steps under our policy ($\pi$ or $\mu$). This is an unnormalized version of the visitation density \cite{sutton2018} of a stochastic process which at every step either proceeds as above with probability $\gamma$, or jumps to an abstract terminal state with probability $1-\gamma$. If an explicit set of terminal states is defined and the process is guaranteed to reach one of them in finite time with probability $1$, the discount factor can be set to $\gamma =1$. This setting is called episodic.

With these preliminaries, existing policy gradient results can be summarized as follows. In the stochastic case we have the original Policy Gradient theorem \cite{sutton1999}:
\begin{equation}\label{GS}
\nabla_{\theta}J_S(\theta) = \mathbb{E}_{x \sim \rho_S(\cdot, \theta), \thickspace u \sim \pi(\cdot|x, \theta)} 
\left[ \nabla_\theta \ln \pi(u|x,\theta) \thickspace Q_S(x, u)) \right]
\end{equation}
In the deterministic case we have the Deterministic Policy Gradient theorem \cite{silver2014}:
\begin{equation}\label{GD}
\nabla_{\theta}J_D(\theta) =  \mathbb{E}_{x \sim \rho_D(\cdot, \theta)} 
\left[ \nabla_\theta \mu(x,\theta) \thinspace \left. \nabla_u Q_D(x, u)) \right|_{u = \mu(x,\theta)} \right]
\end{equation}

\noindent Policy gradient methods can sometimes be accelerated by using a ``natural'' gradient \cite{kakade2001,peters2008} defined as:
\begin{equation}
\nabla_\theta^{\scriptscriptstyle{\textrm{NAT}}} J(\theta) \equiv F(\theta)^{-1} \nabla_\theta J(\theta)
\end{equation}
Here $F$ is an s.p.d. matrix related to Fisher information. In the stochastic case $F$ is given by \cite{kakade2001}:
\begin{equation}\label{FS}
F_S(\theta) = \mathbb{E}_{x \sim \rho_S(\cdot, \theta), \thickspace u \sim \pi(\cdot|x, \theta)} \left[ \nabla_\theta \ln \pi(u|x,\theta) \thickspace \nabla_\theta \ln \pi(u|x,\theta)^T \right]
\end{equation}
In the deterministic case $F$ is given by \cite{silver2014}:
\begin{equation}\label{FD}
F_D(\theta) = \mathbb{E}_{x \sim \rho_D(\cdot, \theta)} \left[ \nabla_\theta \mu(x,\theta) \thickspace \nabla_\theta \mu(x,\theta)^T \right]
\end{equation}

\noindent We seek to understand how the two cases (S) and (D) related.

\subsection{Background notes}

The deterministic result can be obtained from the stochastic result by taking a zero-noise limit, in an elaborate way \cite{silver2014}. A simpler way to recover (\ref{GD}) from (\ref{GS}) is to re-write (\ref{GS}) in a form resembling policy iteration \cite{szepesvari2022,bhandari2024}, as follows. Using the generic identities $\nabla_x g(x) = g(x) \nabla_x \ln g(x)$ and $\nabla_x g(x) = \left. \nabla_y g(x+y)\right|_{y = 0}$, the inner expectation in (\ref{GS}) becomes:
\begin{equation}
\begin{aligned}
\mathbb{E}_{u \sim \pi(\cdot|x, \theta)} 
\left[ \nabla_\theta \ln \pi(u|x,\theta)  Q(x, u,\theta) \right] & = 
\int \nabla_\theta \pi(u|x,\theta)  Q(x, u,\theta) =
\int \left. \nabla_y \pi(u|x,\theta+y) \right|_{y = 0} Q(x, u,\theta) du \\
& = \left. \nabla_y \left( \int \pi(u|x,\theta+y)  Q(x, u,\theta) du \right) \right|_{y = 0}
\end{aligned}
\end{equation}
Changing the order of differentiation and integration requires mild regularity assumptions. Now set $\pi$ to a delta function corresponding to a deterministic policy: $\pi(u|x,\theta) = \delta(u - \mu(x,\theta))$. The above integral then evaluates to $Q(x, \mu(x,\theta + y),\theta)$, and so its gradient (using the chain rule) becomes:
\begin{equation}
\left. \nabla_y Q(x, \mu(x,\theta + y),\theta) \right|_{y = 0} = \nabla_\theta \mu(x,\theta) \left.  \nabla_u Q(x, u, \theta) \right|_{u = \mu(x,\theta)}
\end{equation}
which coincides with the deterministic policy gradient in (\ref{GD}). 

The natural gradient is related to Differential Geometry. Consider a manifold parameterized by $\theta$ and a scalar function $J(\theta)$ on the manifold. The gradient $\nabla J$ is a vector in the tangent space at $\theta$. The vector of partial derivatives $dJ = (\partial J/\partial \theta_i)$ is a differential form which lives in the dual (or co-tangent) space. Given a metric $M(\theta)$ on the manifold, the mapping from co-tangent to tangent vectors is:
\begin{equation}
\nabla J = M^{-1} dJ
\end{equation}
It is common practice outside Differential Geometry to use $dJ$ and call it $\nabla J$, even though the two are only equal under the Cartesian metric $M = I$. For general metrics, the above correction should be applied. What it does is find the steepest ascent direction while measuring direction vector lengths as $v^T M v$. The metric used by natural gradients is Fisher information. It comes from statistical manifolds \cite{amari1998}. A better (but harder to estimate) metric in the context of optimization is the Hessian of $J(\theta)$. If we establish $\nabla_\theta J_S(\theta) = \nabla_\theta J_D(\theta)$ for all $\theta$, then the Hessians and all higher-order derivatives are also equal. In contrast, equality of the gradients does not imply equality of $F_S(\theta)$ and $F_D(\theta)$; this needs to be addressed separately.

\section{Equivalence in a widely-used family of MDPs}

In the general MDP formulation outlined above, we can only see vague similarities between $\nabla_{\theta}J_S$ and $\nabla_{\theta}J_D$, and also between $F_S$ and $F_D$. We know that they are the correct gradients for the corresponding MDPs, but how are they related and are they ever the same? In this section we focus on problems where the cost is quadratic in the control and the control noise is Gaussian, while the dynamics and state cost remain general. In this restricted setting we can utilize special properties of Gaussian densities and closed-form expectations of quadratics over Gaussians, and establish equivalence between (S) and (D).

\subsection{Restriction to Gaussian control noise}

Stochastic policies in continuous control are usually constructed by adding Gaussian noise to the output of an underlying deterministic policy:
\begin{equation}
u = \mu(x, \theta) + d, \quad \textrm{where} \thickspace d \sim \mathcal{N}(0, \Sigma)
\end{equation}
Thus the stochastic policy expressed as a density is $\pi(u|x, \theta) = p_\Sigma (u - \mu(x, \theta))$, where:
\begin{equation}
p_\Sigma(y) \equiv \left| 2 \pi \Sigma \right|^{-1/2} \exp \left( -\tfrac{1}{2} y^T \Sigma^{-1} y \right)
\end{equation}
The common term $(\nabla \ln \pi)$ appearing in (\ref{GS}, \ref{FS}) then becomes:
\begin{equation}
\nabla_\theta \ln \pi(u|x,\theta) = \nabla_\theta \mu(x, \theta) \thinspace \Sigma^{-1}  \thinspace (u - \mu(x, \theta))
\end{equation}
Note that unlike the proof of the Deterministic Policy Gradient theorem \cite{silver2014} which involves a zero-noise limit of a stochastic policy, we will not be taking such a limit here; our $\Sigma$ is fixed.

Dynamics are usually based on a deterministic simulator which computes some function $x' = f(x,u)$. Since most physical systems have (soft or hard) state constraints, adding noise directly to the next state is not a good idea. Instead, it is common practice to leverage the stochasticity of the policy, and rely on the simulator to enforce relevant state constraints even for noisy controls. Thus the transition density is:
\begin{equation}
p_S(x'|x,u) = \delta(x' - f(x,u))
\end{equation}

\noindent Next we turn to the deterministic case, where the policy is $u = \mu(x, \theta)$. Here we can no longer leverage stochasticity coming from the policy, and need some other mechanism for adding noise (the Deterministic Policy Gradient theorem requires proper densities.) This is usually done by adding Gaussian noise $d \sim \mathcal{N}(0, \Sigma)$ to the controls in the dynamics stage, and then applying the deterministic simulator function. This yields (now stochastic) dynamics $x' = f(x, u + d)$. The resulting transition density is complicated, but fortunately we will not need its explicit form here. When the constraints are soft and the mapping between controls and feasible next states is smooth and one-to-one -- as in MuJoCo physics \cite{todorov2014} as well as simulators that implement constraints via spring-dampers -- this density restricted to feasible next states is:
\begin{equation}
p_D(x'|x,u) = \left| \left. \nabla_y f(x,y) \right|_{y = u + d} \right| \thickspace p_\Sigma (d), \quad \textrm{where} \thickspace d = f^{-1}\left(x, x'\right) - u
\end{equation}

\noindent Thus in both problem formulations (S) and (D), the controlled state transitions are:
\begin{equation}
x' = f(x, \mu(x, \theta) + d),\quad \textrm{where} \thickspace d \sim \mathcal{N}(0, \Sigma)
\end{equation}
Problem (S) involves stochastic policy and deterministic dynamics, while problem (D) involves deterministic policy and stochastic dynamics; but that is just a difference in perspective.

There is however a real difference between the two problems, and it comes down to how the cost is evaluated. In (S), noise is already added to the control in the policy stage, and so the cost is evaluated at $u = \mu(x, \theta) + d$. While in (D), noise is added to the control later in the dynamics stage, and so the cost is evaluated at $u = \mu(x, \theta)$.

We are now ready to specialize the relevant quantities from the previous section to the Gaussian control noise setting developed in this section. Since the controlled state transitions in (S) and (D) are the same, the discounted densities are also the same:
\begin{equation}
\rho_S(x, \theta) = \rho_D(x, \theta)
\end{equation}

\noindent The natural gradient correction matrix $F_S$ becomes:
\begin{equation}
\begin{aligned}    
F_S(\theta) & = \mathbb{E}_{x \sim \rho, \thickspace d \sim p_\Sigma} \left[ \nabla_\theta \mu(x, \theta) \Sigma^{-1} d d^T \Sigma^{-1} \nabla_\theta \mu(x, \theta) ^T \right] \\
& = \mathbb{E}_{x \sim \rho} \left[ \nabla_\theta \mu(x, \theta) \thinspace \Sigma^{-1} \thinspace \nabla_\theta \mu(x, \theta) ^T \right]
\end{aligned}    
\end{equation}
while $F_D$ remains as in (\ref{FD}). Comparing the two expressions suggests that $F_D$ should be modified, using $\Sigma^{-1}$ as a metric in control space:
\begin{equation}
\tilde{F}_D(\theta) \equiv \mathbb{E}_{x \sim \rho} \left[ \nabla_\theta \mu(x,\theta) \thinspace 
 \Sigma^{-1} \thinspace \nabla_\theta \mu(x,\theta)^T \right]
\end{equation}
With this modification, the natural gradient correction matrices are the same:
\begin{equation}
F_S(\theta) = \tilde{F}_D(\theta)
\end{equation}

\noindent The remaining quantities turn out to be similar but not identical. The state-control value function becomes:
\begin{equation}\label{Q}
\begin{aligned}
Q_S(x,u) &= \ell(x, u) + \gamma v_S(f(x,u))] \\
Q_D(x,u) &= \ell(x, u) + \gamma \mathbb{E}_{d \sim p_\Sigma} [v_D(f(x,u + d))]
\end{aligned}    
\end{equation}
The Bellman equation for the state value function becomes:
\begin{equation}\label{v}
\begin{aligned}
v_S(x) & = \mathbb{E}_{d \sim p_\Sigma} \left[ \ell(x, \mu(x, \theta) + d)  +
 \gamma v_S(f(x,\mu(x, \theta) + d))) \right] \\
v_D(x) & =  \ell(x, \mu(x, \theta)) + \gamma \mathbb{E}_{d \sim p_\Sigma} \left[
 v_D(f(x,\mu(x, \theta) + d))) \right]
\end{aligned}
\end{equation}

\noindent For the policy gradient, setting $u = \mu(x, \theta) + d$ in the stochastic case, we have:
\begin{equation}
\begin{aligned}
\nabla_{\theta}J_S(\theta) & = \mathbb{E}_{x \sim \rho, \thickspace d \sim p_\Sigma} 
\left[ \nabla_\theta \mu(x, \theta) \thinspace \Sigma^{-1} \thinspace d \thinspace Q_S(x, u)) \right] \\
& = \mathbb{E}_{x \sim \rho} 
\left[ \nabla_\theta \mu(x, \theta) \thickspace  \Sigma^{-1} \thickspace \mathbb{E}_{d \sim p_\Sigma} \left[ \thinspace d \thinspace Q_S(x, u)) \right] \right] \\
& = \mathbb{E}_{x \sim \rho} 
\left[ \nabla_\theta \mu(x, \theta) \thickspace \mathbb{E}_{d \sim p_\Sigma} \left[ \nabla_u Q_S(x, u)) \right] \right]
\end{aligned}    
\end{equation}
The last step follows from Stein's lemma, which states that if $y \sim \mathcal{N}(\mu,\Sigma)$, then for any smooth scalar function $g(y)$ we have:
\begin{equation}
\mathbb{E}_y \left[ (y - \mu) g(y) \right] = \Sigma \thinspace \mathbb{E}_y \left[ \nabla g(y) \right]
\end{equation}
Note that while Stein's lemma appears general, its proof uses integration by parts and relies on the specific formula for a Gaussian density; so it is not valid for general densities.

Thus the policy gradient becomes:
\begin{equation}\label{grad}
\begin{aligned}
\nabla_{\theta}J_S(\theta) &= \mathbb{E}_{x \sim \rho} 
\left[ \nabla_\theta \mu(x, \theta) \thickspace \mathbb{E}_{d \sim p_\Sigma} \left[ \left. \nabla_u Q_S(x, u)) \right|_{u = \mu(x, \theta) + d} \right] \right] \\
\nabla_{\theta}J_D(\theta) &= \mathbb{E}_{x \sim \rho} 
\left[ \nabla_\theta \mu(x,\theta) \thinspace \left. \nabla_u Q_D(x, u)) \right|_{u = \mu(x,\theta)} \right]
\end{aligned}    
\end{equation}

\noindent The expressions for (S) and (D) in (\ref{Q}, \ref{v}, \ref{grad}) are quite similar. The only difference within each pair is that $d \sim p_\Sigma$ in one expression corresponds to $d = 0$ in the other. We can make the noise covariance $\Sigma(x)$ state-dependent throughout this section, and all results still hold.

\subsection{Further restriction to quadratic control cost}

We now focus on problems where the cost can be arbitrary in the state but must be quadratic in the control:
\begin{equation}
\ell(x, u) = q(x) + u^T r(x) + u^T R u 
\end{equation}
where $R$ is symmetric. This construction is quite natural; indeed quadratics are the most commonly used form of control cost in the literature.

Using formulas for the expectation of a quadratic under a Gaussian, we have:
\begin{equation}
\begin{aligned}
\mathbb{E}_{d \sim p_\Sigma} \left[ \ell(x, \mu + d) \right] 
& = q(x) + \mathbb{E}_{d \sim p_\Sigma} \left[(\mu + d)^Tr(x) +  (\mu + d)^T R (\mu + d) \right] \\
& = q(x) + \mu^T r(x) + \mu^T R \mu + \textrm{tr} (R\Sigma) \\
& = \ell(x, \mu) + \textrm{tr} (R\Sigma)
\end{aligned}    
\end{equation}
Since the costs turned out to be identical up to a constant, the state value functions (which are cumulative costs) are also identical up to a constant:
\begin{equation}
v_S(x,\theta) = v_D(x,\theta) + c, \quad \textrm{where} \thickspace c =  (1-\gamma)^{-1} \textrm{tr} (R\Sigma)
\end{equation}

\noindent The state-control value functions however are still different. In particular, $Q_D$ contains an extra convolution with respect to the noise density, making it somehow smoother than $Q_S$:
\begin{equation}\label{Qdif}
Q_D(x) - Q_S(x)  = \gamma \mathbb{E}_{d \sim p_\Sigma} \left[ v_D(f(x,\mu(x,\theta)+d)) \right] - \gamma v_D(f(x,\mu(x,\theta)) - \gamma c
\end{equation}
It seems the only way to make $Q_S$ and $Q_D$ identical (up to a constant) would be to adopt the LQG problem formulation -- where $v$ is quadratic and $f$ is linear. But LQG is too restrictive.

Even though $Q_S$ and $Q_D$ turned out to be different, their difference is exactly canceled by the difference in the expressions for $\nabla J_S$ and $\nabla J_D$, making the policy gradients themselves identical. To show this, we express $Q$ in terms of $v$ and substitute in the formulas for $\nabla J$. The terms that are different between the two expressions are denoted $T$. For $T_S$ we have:
\begin{equation}
\begin{aligned}
T_S(x) & \equiv \mathbb{E}_{d \sim p_\Sigma} \left[ \left. \nabla_u Q_S(x, u)) \right|_{u = \mu(x, \theta) + d} \right] \\
& = \mathbb{E}_{d \sim p_\Sigma} \left[ \left. \left( r(x) + 2Ru + \gamma \nabla_u v_S(f(x,u)) \right) \right|_{u = \mu(x, \theta) + d} \right] \\
& = r(x) + 2 R \mu(x,\theta) + \gamma \mathbb{E}_{d \sim p_\Sigma} \left[ \left. \nabla_u v_S(f(x,u)) \right|_{u = \mu(x, \theta) + d} \right]
\end{aligned}    
\end{equation}
For $T_D$ we have:
\begin{equation}
\begin{aligned}
T_D(x) & \equiv \left. \nabla_u Q_D(x, u)) \right|_{u = \mu(x,\theta)} \\
& = r(x) + 2 R \mu(x,\theta) + \gamma \mathbb{E}_{d \sim p_\Sigma} \left[ \left. \nabla_u v_D(f(x,u + d)) \right|_{u = \mu(x, \theta)}\right]
\end{aligned}    
\end{equation}
Now $v_S$ and $v_D$ only differ by a constant, eliminated by the differentiation. The generic identity:
\begin{equation}
\left. \nabla_u g(u) \right|_{u = \mu + d} = \left. \nabla_u g(u+d) \right|_{u = \mu}
\end{equation}
then implies that $T_S(x) = T_D(x)$, and therefore
\begin{equation}
\nabla_\theta J_S(\theta) = \nabla_\theta J_D(\theta)
\end{equation}

\noindent We could make the control cost Hessian state-dependent, but in that case we also have to make the noise covariance state-dependent, so that $R(x) = \alpha \Sigma(x)^{-1}$ for some constant $\alpha$. Then $\textrm{tr}\left(R(x)\Sigma(x)\right) = \textrm{tr}(\alpha I)$ and our results still hold. If this term is not constant, the value functions $v_S(x)$ and $v_D(x)$ differ by some unknown function of $x$, and we can no longer show equivalence.

\subsection{Summary of equivalence results for Quadratic-Gaussian MDPs}

Here we defined stochastic (S) and deterministic (D) policies, along with corresponding dynamics and costs in such a way that:
\begin{enumerate}

\item The stochastic state transitions under both policies are identical:
\begin{equation}
x' = f(x, \mu(x, \theta) + d),\quad \textrm{where} \thickspace d \sim \mathcal{N}(0, \Sigma)
\end{equation}
even though the noise $d$ originates from the policy in (S), and from the dynamics in (D).

\item The cost function in both cases is:
\begin{equation}
\ell(x,u) = q(x) + u^T r(x) + u^T R u
\end{equation}
however it is evaluated at $u = \mu(x, \theta) + d$ in (S), and at $u = \mu(x, \theta)$ in (D).

\item The discounted densities, natural gradient correction matrices, and policy gradients are identical:
\begin{equation}
\begin{aligned}
\rho_S(x, \theta) & = \rho_D(x, \theta) \\
F_S(\theta) & = \tilde{F}_D(\theta) \\
\nabla_\theta J_S(\theta) & = \nabla_\theta J_D(\theta)
\end{aligned}    
\end{equation}

\item The state value functions are identical up to a constant:
\begin{equation}
v_S(x) = v_D(x) + c, \quad \textrm{where} \thickspace c = (1-\gamma)^{-1} \textrm{tr}(R\Sigma)
\end{equation}

\item The state-control value functions $Q_S(x)$ and $Q_D(x)$ are different in a non-trivial way (\ref{Qdif}).

\item The noise covariance and control cost Hessian can be made state-dependent, and the results still hold as long as $R(x) = \alpha \Sigma(x)^{-1}$ for some constant $\alpha$.

\end{enumerate}

\noindent This equivalence between stochastic and deterministic policy gradients relies on special properties of Gaussian control noise and quadratic control costs. Nevertheless, this special case appears to cover many (if not most) applications of policy gradient methods in continuous control.

\subsection{Learning algorithms}

\noindent Putting everything together, the exact policy gradient in both our formulations (S) and (D) can be obtained by solving the Bellman equation:
\begin{equation}\label{Bel}
v(x) = \ell(x, \mu(x, \theta)) + \gamma \mathbb{E}_{u \sim \mathcal{N}(\mu(x,\theta), \Sigma)} \left[ v(f(x,u)) \right]
\end{equation}
and then evaluating:
\begin{equation}\label{gradFinal}
\nabla_\theta J(\theta) = \mathbb{E}_{x \sim \rho(\cdot, \theta)} 
\left[ \nabla_\theta \ell(x, \mu(x, \theta)) + \nabla_\theta \mu(x, \theta) \gamma \mathbb{E}_{u \sim \mathcal{N}(\mu(x,\theta), \Sigma)} \left[ K(x,u) \right] \right)
\end{equation}
where $K(x,u)$ can be defined in two ways, both of which yield the same policy gradient in expectation:
\begin{equation}\label{K}
\begin{aligned}
K_S(x,u) & \equiv \Sigma^{-1} (u - \mu(x,\theta)) v(f(x,u)) \\
K_D(x,u) & \equiv \nabla_u v(f(x,u))
\end{aligned}    
\end{equation}

\noindent Learning algorithms derived from (\ref{Bel}, \ref{gradFinal}, \ref{K}) would be model-based, requiring access to $f$ and $\ell$. The true $v$ would be replaced with empirical sums of costs, or with a function approximation $\hat{v}(x, \omega)$ learned by minimizing Bellman residuals with respect to $\omega$ for fixed $\theta$. The expectations would be replaced with stochastic approximations along sample trajectories. The control cost here is evaluated at the deterministic control, reducing the variance of the gradient estimate. Learning $v(x)$ rather than $Q(x,u)$ should further reduce overall approximation error, due to the lower dimensionality. 

The variance of stochastic approximation schemes can often be reduced by subtracting a suitable baseline term that leaves the expectations unchanged. This is common in advantage estimation. It can also be done here, in the absence of advantages and $Q$ functions, by modifying $K_S, K_D$ as:
\begin{equation}\label{baseline}
\begin{aligned}
\tilde{K}_S(x,u) & \equiv \Sigma^{-1} (u - \mu) \left( v(f(x,u)) - \hat{v}(f(x,\mu)) \right) \\
\tilde{K}_D(x,u) & \equiv \nabla_u v(f(x,u)) - \nabla_{\mu\mu} \hat{v}(f(x,\mu)) \thickspace (u - \mu)
\end{aligned}    
\end{equation}
perhaps using the Gauss-Newton approximation $\nabla_\mu f \thinspace \nabla_{ff} \hat{v} \thinspace \nabla_\mu f^T$ in place of the Hessian $\nabla_{\mu\mu} \hat{v}$.

Using $K_D$ requires not only a function approximation $\hat{v}(x, \omega)$ but also the control derivative $\nabla_u f(x,u)$. This derivative can be computed analytically in MuJoCo, and will be available in our upcoming Optico toolbox for model-based control. In future work, we look forward to implementing and testing new algorithms based on the present theoretical results.

\section{General construction of pairs of equivalent MDPs}

The above equivalence was obtained for the same MDP under two policies: stochastic and deterministic. Here we ask a different question. Given any MDP with stochastic policy, can we construct a corresponding MDP with deterministic policy, such that the two (now different) MDPs have the same value function and controlled state transitions -- and therefore the same policy performance and policy gradient? This indeed turns out to be possible. We state the formal result later, but first we provide the intuition.

The idea is to ``look under the hood'' of the stochastic policy $\pi(u|x,\theta)$ and ask how exactly it generates controls. We break it down into a deterministic stage that extracts some minimal set of density parameters $\eta = \mu(x,\theta)$, which are then passed to some density $\tilde{\pi}(u|\eta)$ that generates the stochastic controls:
\begin{equation}
(x, \theta) \longrightarrow \eta = \mu(x, \theta) \longrightarrow u \sim \tilde{\pi}(\cdot|\eta)
\end{equation}
The intermediate quantity $\eta$ is closely related to the concept of sufficient statistics. Does such an $\eta$ always exist? It does, but in the worst case it could be just $\eta = (x, \theta)$. This is undesirable when $\theta$ is high-dimensional (such as the weights of a neural network policy.) In the previous section, $\eta$ was the mean of the Gaussian from which controls were sampled, so it had the same dimensionality as $u$. Anything lower-dimensional would imply that some projection of the controls is independent of the states. This actually happens when dimensionality-reducing control synergies are used: $\eta$ become the synergy activations and $\dim(\eta) < \dim(u)$.

The new MDP we construct below will have the same state $x$ and policy parameters $\theta$ as the original MDP, but the control will be $\eta$ rather than $u$. Its deterministic policy will be $\eta = \mu(x,\theta)$ corresponding to the first stage of the stochastic policy in the original MDP. We will then define the cost and transition probability of the new MDP so as to make it equivalent to the original MDP.

\begin{theorem} Let  $x \in \mathcal{R}^{n_x}, u \in \mathcal{R}^{n_u}, \theta \in \mathcal{R}^{n_\theta}, \eta \in \mathcal{R}^{n_\eta}$ be real-valued vectors.
Let S-MDP denote the MDP with state $x$, control $u$, step cost $\ell_S(x,u)$ and transition probability $p_S(x'|x,u)$. Let $\pi_S(u|x,\theta)$ be a parametric stochastic policy in the form:
\begin{equation}\label{pitilde}
\pi_S(u|x,\theta) = \tilde{\pi}\left(u|\mu(x,\theta)\right)
\end{equation}
for some density $\tilde{\pi}\left(u|\eta\right)$ and function $\eta = \mu(x,\theta)$. Let D-MDP denote the MDP with state $x$, control $\eta$, and step cost $\ell_D(x,\eta)$ and transition probability $p_D(x'|x,\eta)$ defined as:
\begin{equation}\label{Ddef}
\begin{aligned}
\ell_D(x,\eta) & \equiv \int \tilde{\pi}(u| \eta) \ell_S(x,u) du \\
p_D(x'|x,\eta) & \equiv \int \tilde{\pi}(u| \eta) p_S(x'|x,u) du
\end{aligned}
\end{equation}
Let $v_S(x,\theta)$ denote the state value function (episodic or average) under stochastic policy $\pi_S(u|x,\theta)$ in the S-MDP. Let $v_D(x,\theta)$ denote the state value function under deterministic policy $\eta = \mu(x,\theta)$ in the D-MDP. Let $P_S(x'|x,\theta)$ and $P_D(x'|x,\theta)$ denote the controlled state transition probabilities under the corresponding policies in the S-MDP and D-MDP. Then for all $x',x,\theta$,
\begin{equation}
\begin{aligned}  
P_S(x'|x,\theta) & = P_D(x'|x,\theta) \\
v_S(x,\theta) & = v_D(x,\theta)
\end{aligned}
\end{equation}
\end{theorem}

\noindent \textbf{Proof.} The controlled state transition probabilities in the S-MDP and D-MDP are:
\begin{equation}
\begin{aligned}   
P_S(x'|x,\theta) & = \int \pi(u|x,\theta) p_S(x'|x,u) du \\
P_D(x'|x,\theta) & = p_D(x'|x, \mu(x,\theta))
\end{aligned}
\end{equation}
To show that they are equal, substitute (\ref{pitilde}) in the above expression for $P_S$, yielding:
\begin{equation}
P_S(x'|x,\theta) = \int \tilde{\pi}(u|\mu(x,\theta)) p_S(x'|x,u) du = p_D(x'|x, \mu(x,\theta)) = P_D(x'|x,\theta)
\end{equation}
For the value functions, we present the proof in the $\gamma$-discounted infinite-horizon setting (other settings are similar.) The policy-specific value function is known to satisfy the Bellman equation. For the S-MDP and D-MDP as defined in the theorem, this equation is:
\begin{equation}\label{SDBel}
\begin{aligned}   
v_S(x,\theta) & = \int \pi(u|x, \theta) \left( \ell_S(x,u) + \gamma \int p_S(x'|x,u) v_S(x', \theta) dx' \right) du \\
v_D(x,\theta) & = \ell_D(x,\mu(x,\theta)) + \gamma \int p_D(x'|x,\mu(x,\theta)) v_D(x', \theta) dx'
\end{aligned}
\end{equation}
We now substitute (\ref{pitilde}) in the above Bellman equation for the S-MDP, yielding:
\begin{equation}
\begin{aligned}   
v_S(x,\theta) & = \int \tilde{\pi}(u|\mu(x,\theta)) \left( \ell_S(x,u) + \gamma \int p_S(x'|x,u) v_S(x', \theta) dx' \right) du \\
& = \underbrace{\int \tilde{\pi}(u|\mu(x,\theta)) \ell_S(x,u) du}_{\ell_D(x,\mu(x,\theta))} + \gamma \int  \underbrace{\left( \int \tilde{\pi}(u|\mu(x,\theta)) p_S(x'|x,u) du \right)}_{p_D(x'|x,\mu(x,\theta))} v_S(x',\theta) \thinspace dx' 
\end{aligned}
\end{equation}
where we have recognized the definitions (\ref{Ddef}). The last expression is identical to the Bellman equation for $v_D$ in (\ref{SDBel}). Since the Bellman equation has a unique solution, $v_S(x,\theta) = v_D(x,\theta)$.
$\square$

\begin{corollary} The equality of the controlled transition probabilities implies that if both S-MDP and D-MDP have the same initial state distributions $p_0(x)$, the visitation densities and stationary densities (when the latter exists) are the same. And since the value functions are the same, the policy performance $J(\theta)$ and policy gradients $\nabla_\theta J(\theta)$ are also the same.
\end{corollary}

\noindent As in the special case studied earlier, the S-MDP cost here is evaluated at a random control while the D-MDP cost is evaluated at the deterministic control. How is it then that we end up with the same value function? This is possible because the value function involves expectations.

The practical implication of this general result is as follows. Any MDP with stochastic policy is amenable to the stochastic policy gradient theorem. Using our general procedure, we can construct an equivalent MDP with deterministic policy which is amenable to the deterministic policy gradient theorem. While the policy gradient is guaranteed to be the same in expectation, stochastic approximation schemes may behave differently, and which class of methods yields better performance in any given problem is an empirical matter.

\subsection{Application to the Quadratic-Gaussian special case}

We now apply this general result to the special case from the previous section, and see if we recover the earlier equivalence. Following the procedure in the theorem, we start with the S-MDP and construct a corresponding D-MDP. The step cost, transition probability and stochastic policy in this S-MDP are:
\begin{equation}
\begin{aligned}
\ell_S(x,u) & = q(x) + u^T r(x) + u^T R u \\
p_S(x'|x,u) & = \delta(x' - f(x,u)) \\
\pi(u|x,\theta) & = \mathcal{N}(u; \mu(x,\theta), \Sigma)
\end{aligned}
\end{equation}
Thus the minimal density parameters $\eta = \mu(x,\theta)$ here are just the mean of the stochastic policy, and we have $\tilde{\pi}(u|\eta) = \mathcal{N}(u;\eta, \Sigma)$. Note that the constant covariance $\Sigma$ is baked into $\tilde{\pi}$. State and/or parameter-dependent covariance $\Sigma(x,\theta)$ would have to be included in $\eta$.

Applying (\ref{Ddef}) to this S-MDP now yields a corresponding D-MDP with deterministic policy $\eta = \mu(x,\theta)$, and step cost and transition probability:
\begin{equation}
\begin{aligned}
\ell_D(x,\eta) & = q(x) + \eta^T r(x) + \eta^T R \eta + \textrm{tr}(R\Sigma) \\
p_D(x'|x,\eta) & = \left\{ x' = f(x, u), \thickspace u \sim \mathcal{N}(\eta, \Sigma) \right\} 
\end{aligned}    
\end{equation}
This is the same as the special-case D-MDP. The only difference is that here the constant $\textrm{tr}(R\Sigma)$ was added to the deterministic cost, resulting in equal value functions (from the theorem.) While previously we did not add this constant correction to the cost, and so it showed up as a constant offset in the value function.

Are there other examples where the quantities (\ref{Ddef}) needed to construct the equivalent D-MDP can be computed in closed form? We could generalize $\tilde{\pi}$ to a mixture of Gaussians. In principle we could also generalize $p_S$ to a proper Gaussian instead of a delta function, but that would mean adding noise directly to the next state (instead of pushing noise through some deterministic dynamics) and this may not make physical sense. It is likely that additional examples can be constructed.

\section{Discussion}

Here we established equivalence of stochastic and deterministic policy gradients, first in the special case of Quadratic-Gaussian MDPs, and then in a completely general setting.

Recent generalizations of policy gradients have shown that the classic theorems for both stochastic (\ref{GS}) and deterministic (\ref{GD}) policies can be recovered from a single quantity; related to some of our results in the Quadratic-Gaussian special case. One such generalization is Expected Policy Gradient \cite{ciosek2020}. That paper also considered Gaussian policies but assumed that the state-control value function is quadratic in the control; which may be only satisfied in the LQG formulation. Another such generalization, developed in parallel with the present work, is Wasserstein Policy Optimization \cite{pfau2025}. For Gaussian policies assumed to be independent of the state, \cite{pfau2025} showed that their descent direction coincides with the stochastic policy gradient, and resembles the deterministic policy gradient as in (\ref{grad}) here. The unusual assumptions in the above papers were needed to establish the corresponding analytical results. Without such assumptions, both generalizations yield useful descent directions and learning algorithms, unrelated to the topic of equivalence.

Stochastic and deterministic policy gradients are presently seen as having different numerical behaviors, with different strengths and weaknesses. This may be partly because function approximation (in model-free Reinforcement Learning) is typically used to learn the state-control value function, which we showed is the only quantity that is different between the stochastic and deterministic formulations. The implication is that learning the state value function should yield common policy updates. This is true in expectation as shown here, although stochastic approximation schemes may still behave differently.

The general result in Section 4 is itself a special case of an even more general framework called Dynamical System Optimization \cite{todorov2025}. In that framework we established an additional equivalence to Linearly-solvable MDPs \cite{todorov2006}, and obtained policy gradients for multiple MDP families from a single unifying quantity.

\subsection*{Acknowledgements} Thanks to Yuval Tassa, Csaba Szepesvári and David Pfau for their comments on the manuscript.

\end{document}